\documentclass[journal]{IEEEtran}

\usepackage{cite}
\usepackage{amsmath,amssymb,amsfonts}
\usepackage{algorithm}
\usepackage{algpseudocode}
\usepackage{booktabs}
\usepackage{graphicx}
\usepackage{xcolor}
\usepackage{url}

\begin{document}

\title{Agentic Adversarial Rewriting Exposes Architectural Vulnerabilities in \\Black-Box NLP Pipelines
}

\author{Mazal~Bethany,
Kim-Kwang Raymond Choo,~\IEEEmembership{Senior 
Member,~IEEE},
Nishant Vishwamitra, 
and~Peyman~Najafirad,~\IEEEmembership{Senior 
Member,~IEEE}%
\thanks{M.~Bethany, N.~Vishwamitra, K.-K. R.~Choo, and P.~Najafirad are with the Department of Information Systems and Cybersecurity, University of Texas at San Antonio, San Antonio, TX 78249 USA (e-mail: mazal.bethany@utsa.edu; nishant.vishwamitra@utsa.edu; raymond.choo@fulbrightmail.org; peyman.najafirad@utsa.edu).}%
}

\markboth{IEEE Transactions on Information Forensics and Security}%
{Bethany \MakeLowercase{\textit{et al.}}: Agentic Adversarial Rewriting Under Hard-Label Constraints}

\maketitle

\begin{abstract}
Multi-component natural language processing (NLP) pipelines are increasingly deployed for high-stakes decisions, yet no existing adversarial method can test their robustness under realistic conditions: binary-only feedback, no gradient access, and strict query budgets. We formalize this strict black-box threat model and propose a two-agent evasion framework operating in a semantic perturbation space. An Attacker Agent generates meaning-preserving rewrites while a Prompt Optimization Agent refines the attack strategy using only binary decision feedback within a 10-query budget. Evaluated against four evidence-based misinformation detection pipelines, the framework achieves evasion rates of 19.95 to 40.34\% on modern large language model (LLM) based systems, compared to at most 3.90\% for token-level perturbation baselines that rely on surrogate models because they cannot operate under our threat model. A legacy system relying on static lexical retrieval exhibits near-total vulnerability (97.02\%), establishing a lower bound that exposes how architectural choices govern the attack surface. Evasion effectiveness is associated with three architectural properties: evidence retrieval mechanism, retrieval-inference coupling, and baseline classification accuracy. The iterative prompt optimization yields the largest marginal gains against the most robust targets, confirming that adaptive strategy discovery is essential when evasion is non-trivial. Analysis of successful rewrites reveals four exploitation patterns, each targeting failures at distinct pipeline stages. A pattern-informed defense reduces the evasion rate by up to 65.18\%.
\end{abstract}


\section{Introduction}
\label{sec:introduction}

Advances in large language models (LLMs) have given rise to a new class of composite software systems known as \emph{NLP pipelines}, which are multi-component architectures in which natural language serves as the interface at every processing stage: input intake, intermediate reasoning, and output generation. Retrieval-augmented generation (RAG) pipelines~\cite{lewis2020retrieval}, tool-augmented LLM agents~\cite{schick2023toolformer}, multi-agent reasoning frameworks~\cite{yao2023react}, and evidence-based fact-checking systems~\cite{tian2024web} all instantiate this pattern. Unlike monolithic classifiers that expose a single decision boundary, these systems accept free-form natural language at every inter-component boundary, creating a combinatorially vast and semantically rich attack surface. We focus on evidence-based misinformation detection as a representative domain: these systems exhibit a canonical three-stage structure (natural language input, natural language evidence retrieval, and natural language inferential comparison) and are deployed at scale for high-stakes decisions.

Despite their growing deployment, the adversarial vulnerability of multi-stage NLP systems has not been systematically studied. Existing adversarial text attacks rely on token-level substitutions, such as synonym replacement~\cite{jin2020bert} and character-level perturbations~\cite{gao2018black}, guided by gradient or logit-based optimization. These approaches were designed for monolithic classifiers and face fundamental limitations against multi-component pipelines for three reasons: (1) an effective adversarial input must simultaneously disrupt multiple pipeline stages, specifically both retrieval \emph{and} inference, rather than a single decision boundary; (2) real-world deployments expose only binary decisions with no gradient or probability access; and (3) production systems enforce strict query budgets, rate limiting, and per-call costs, ruling out the extensive querying that token-level methods require. A fourth challenge, which we investigate empirically in this work, is that different pipeline architectures may be vulnerable at different processing stages, potentially requiring adaptive, target-specific exploitation rather than a fixed perturbation scheme.

Prior domain-specific approaches have not closed this gap. Du et al.~\cite{du2022synthetic} demonstrated vulnerability in fact-verification systems but assumed write access to the evidence database. XARELLO~\cite{przybyla2024know} employs reinforcement learning but requires prediction probabilities. TREPAT~\cite{przybyla2024attacking} uses LLMs for initial rewritings via beam search, yet still depends on logits. No existing method can attack multi-component pipelines under true black-box conditions, namely binary feedback only with a strict query budget.

\begin{figure*}[!t]
\centering
\includegraphics[width=0.95\textwidth]{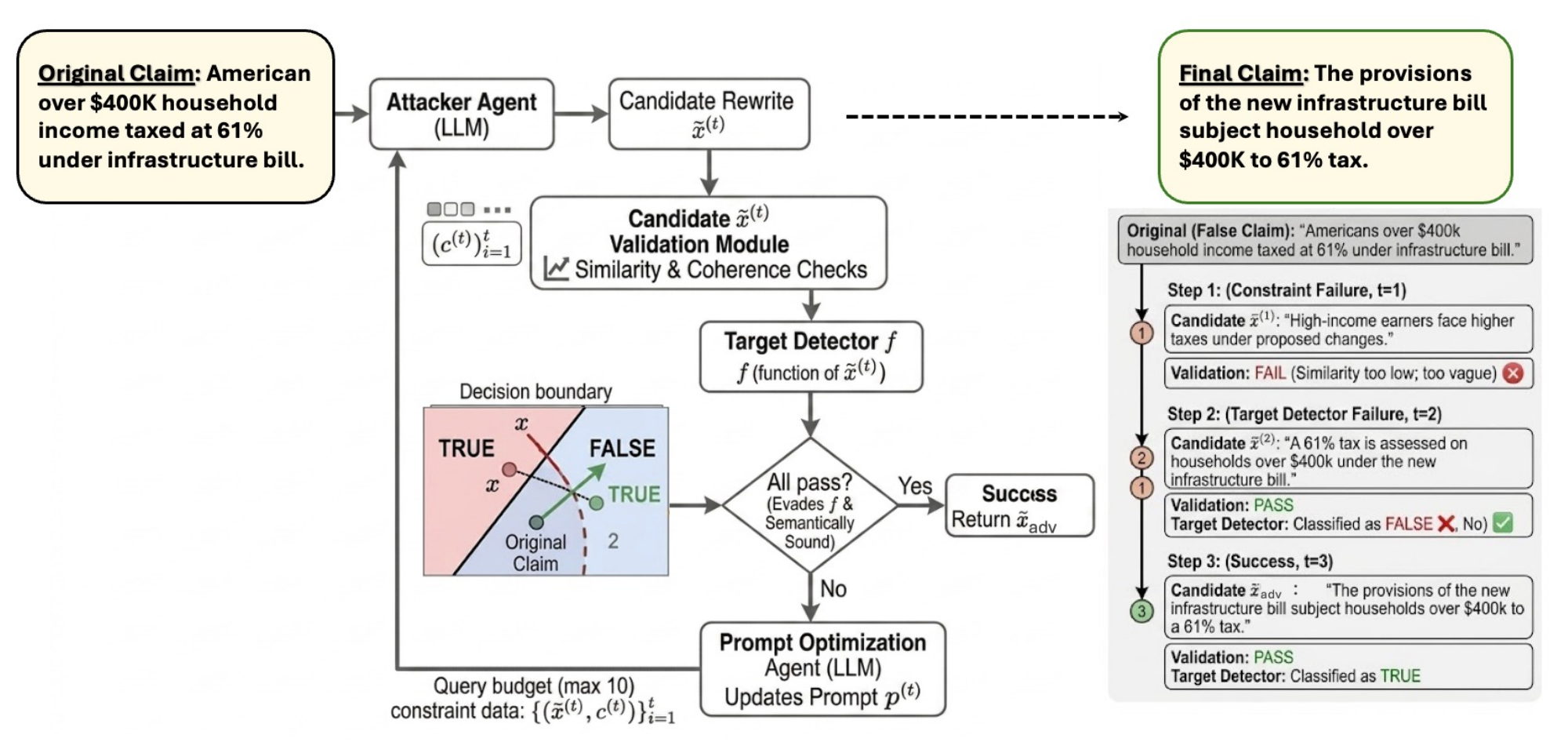}
\caption{High-level overview of the agentic adversarial rewriting framework with a worked example. \textbf{Left:} Iterative attack loop in which the Attacker Agent generates candidate rewrites, the constraint validation module enforces semantic equivalence and linguistic coherence, and the target detector~$f$ provides binary feedback. Failed attempts feed constraint data to the Prompt Optimization Agent, which refines the attack strategy within a 10-query budget. The decision boundary visualization illustrates how successive rewrites navigate the semantic perturbation space. \textbf{Right:} A concrete three-iteration example on a political claim, progressing from a constraint failure (insufficient similarity) through a target detector failure (correct classification) to a successful evasion (misclassified as TRUE).}
\label{fig:overview}
\end{figure*}

We present an agentic LLM framework that exploits black-box NLP pipelines through adversarial rewriting under budget-constrained queries (Fig.~\ref{fig:overview}). A two-agent architecture operates entirely at inference time: an Attacker Agent generates semantically equivalent rewrites through structural and stylistic transformations, while a Prompt Optimization Agent iteratively refines the Attacker's instructions based on binary evaluation feedback. The framework requires no fine-tuning, no gradient access, and operates within a budget of just 10 queries per input. We formalize this as constrained optimization over natural language input space under black-box access.

Evaluated against four evidence-based misinformation detection pipelines, our framework achieves 19.95 to 40.34\% attack success on modern LLM-based systems, compared to at most 3.90\% for token-level baselines that rely on surrogate models. Even the non-iterative Attacker Only variant achieves 14.55 to 35.92\%, confirming that semantic-level rewriting provides a fundamental advantage independent of the optimization strategy. A legacy system with static lexical retrieval is near-completely exploitable (97.02\%), establishing an architectural lower bound. Iterative prompt optimization provides the largest marginal gains against the most robust targets (+10.24\,pp on ICL), and analysis of successful rewrites reveals four exploitation patterns targeting distinct pipeline stages. A pattern-informed defense reduces attack success by up to 65.18\%.

Our contributions are as follows:
\begin{itemize}
\item \textbf{Attack Formulation.} We formalize the problem of exploiting black-box NLP pipelines under budget-constrained queries as constrained optimization over natural language input space, a threat model under which existing token-level methods, constrained to surrogate-model access, achieve at most 3.90\% success on modern pipelines (Section~\ref{sec:formulation}).
\item \textbf{Agentic Attack Framework.} We propose a two-agent framework that generates adversarial rewrites requiring no fine-tuning, no gradient access, and only binary feedback within a 10-query budget (Section~\ref{sec:methodology}).
\item \textbf{Vulnerability Spectrum and Defense.} We evaluate against four pipelines spanning legacy to state-of-the-art architectures, revealing that attack success on modern systems ranges from 19.95\% to 40.34\% ($\pm$4.3\,pp, 95\% CI) and correlates with architectural properties, specifically the evidence retrieval mechanism, retrieval-inference coupling, and baseline accuracy. Analysis uncovers four exploitation patterns associated with distinct pipeline stages, and a pattern-informed defense reduces attack success by up to 65.18\% (Sections~\ref{sec:experiments} through~\ref{sec:defense}).
\end{itemize}

In the next section, we will discuss the extant literature.

\section{Related Work}

\subsection{Multi-Component NLP Systems}
\label{sec:nlp_pipelines}

Modern NLP increasingly relies on multi-component systems where LLMs communicate through natural language at each stage. RAG pipelines~\cite{lewis2020retrieval}, tool-augmented agents~\cite{schick2023toolformer}, reasoning frameworks such as ReAct~\cite{yao2023react}, and compound AI systems~\cite{zaharia2024compound} all instantiate this pattern, creating a semantically rich attack surface that resists adversarial methods designed for single-model classifiers.

Evidence-based misinformation detection is a prominent instance of this multi-stage pattern. Automated approaches span style-based, propagation-based, source-based, and evidence-based methods~\cite{zhou2020survey}. Evidence-based systems produce interpretable decisions grounded in retrieved evidence through a canonical three-stage structure: evidence retrieval, knowledge sourcing, and inferential comparison~\cite{tian2024web,singhal2024evidence}. We select this domain because its multi-stage structure, combined with academic prototypes and commercial APIs, provides a rigorous adversarial testbed.

\subsection{Adversarial Attacks on NLP Systems}

Adversarial attacks on text face fundamentally different challenges than their vision counterparts: text is discrete, and even single-word changes can alter meaning~\cite{alzantot2018generating}. Methods are organized along the white-box to black-box spectrum~\cite{maheshwary2021strong}. Score-based black-box methods receive prediction probabilities and can efficiently identify influential tokens: BAE~\cite{garg2020bae} uses BERT masked language modeling for contextual replacements, BERT-Attack~\cite{li2020bertattack} turns BERT against its own fine-tuned classifiers, and SemAttack~\cite{wang2022semattack} optimizes perturbations across multiple semantic spaces. Decision-based (hard-label) methods receive only the final label, creating a substantially harder optimization problem~\cite{ye2022leapattack}.

Established score-based methods including CLARE~\cite{li2021contextualized}, DeepWordBug~\cite{gao2018black}, TextBugger~\cite{li2019textbugger}, and TextFooler~\cite{jin2020bert} were designed for monolithic classifiers. The TextAttack framework~\cite{morris2020textattack} provides a unified evaluation platform for these methods. As we demonstrate in Section~\ref{sec:experiments}, all achieve at most 3.90\% evasion against modern LLM-based pipelines because they cannot reason about multi-stage interactions. This comparison is structurally asymmetric: these baselines must rely on surrogate models because they require prediction probabilities unavailable from the target. We include them as the best available methods and interpret the gap as evidence that multi-stage pipeline attacks require fundamentally different approaches.

\subsection{Hard-Label Adversarial Attacks on Text}
\label{sec:hardlabel}

Hard-label attacks operate under the most restrictive black-box threat model, where the adversary observes only the final classification label. This constraint transforms adversarial search from continuous optimization into combinatorial search over discrete text, making query efficiency the central design objective.

Maheshwary et al.~\cite{maheshwary2021hardlabel} established the first hard-label text attack using genetic optimization. Subsequent work improved query efficiency through gradient estimation~\cite{ye2022leapattack}, embedding-space optimization~\cite{ye2022texthoaxer}, geometry-aware perturbation~\cite{ye2023pat}, and perturbation minimization~\cite{ye2023hqaattack}. More recent methods further reduce query cost via LIME-based importance ranking~\cite{zhu2024limeattack}, dual-gradient fusion~\cite{qiu2025qeattack}, and hybrid optimization~\cite{liu2024hygloadattack,hu2024fasttextdodger,peng2024textcheater,liu2023sspattack}.

Despite this progress, all existing hard-label methods share two fundamental limitations: they operate at the \emph{token level}, modifying individual words through synonym substitution, and they target \emph{single-model} architectures. Neither property suffices against multi-component pipelines where the adversarial input must simultaneously disrupt retrieval \emph{and} inference. Our framework operates at the \emph{semantic level}, generating holistic sentence-level rewrites, and is the first to demonstrate effective hard-label attacks against multi-stage NLP pipelines.

\subsection{LLM-Based Adversarial Attacks}
\label{sec:llm_attacks}

A growing body of work leverages LLMs as adversarial tools. Perez et al.~\cite{perez2022redteaming} pioneered automated red-teaming by using one LLM to generate test cases that expose harmful behaviors in a target LLM. Zou et al.~\cite{zou2023universal} proposed GCG, which appends optimized adversarial suffixes achieving high evasion on both open-source and closed-source models. PromptAttack~\cite{xu2024promptattack} composes structured attack prompts at multiple granularities, increasing evasion by 42\% using only black-box queries. BEAST~\cite{sadasivan2024beast} achieves gradient-free jailbreaking in under one minute with transferable adversarial suffixes.

Our framework shares the use of LLMs as the attack engine but differs in three critical aspects: (1)~we target multi-component pipelines rather than individual LLMs, (2)~we enforce strict semantic equivalence and coherence constraints, and (3)~we operate under a budget of only 10 queries, whereas GCG and similar methods require thousands of optimization steps.

\subsection{Adversarial Attacks on Pipeline and RAG Systems}
\label{sec:pipeline_attacks}

Recent work has examined the adversarial vulnerability of multi-component systems. Greshake et al.~\cite{greshake2023indirect} demonstrated indirect prompt injection in LLM-integrated applications. PoisonedRAG~\cite{zou2024poisonedrag} achieves 90\% evasion by injecting five malicious documents into a knowledge database, while Phantom~\cite{chaudhari2024phantom} and TrojanRAG~\cite{cheng2024trojanrag} mount backdoor attacks through poisoned retrieval corpora.

In misinformation detection, Du et al.~\cite{du2022synthetic} demonstrated vulnerability through synthetic evidence injection but assumed write access to the evidence database. Abdelnabi and Fritz~\cite{abdelnabi2023factsaboteurs} proposed a taxonomy of evidence manipulation attacks, and Thorne and Vlachos~\cite{thorne2019adversarial} evaluated adversarial attacks on the FEVER shared task. XARELLO~\cite{przybyla2024know} employs reinforcement learning and TREPAT~\cite{przybyla2024attacking} uses LLM-based beam search, but both require prediction probabilities or extensive querying. Our framework is the first to attack multi-component fact-checking pipelines under the most restrictive hard-label threat model with a strict 10-query budget, operating entirely through input-side adversarial rewriting.

\subsection{Defenses Against Adversarial Text Attacks}
\label{sec:defense_related}

Defenses span three paradigms~\cite{goyal2023survey}: certified robustness through randomized smoothing~\cite{ye2020safer,zeng2023certified}, input purification through masked language model infilling~\cite{li2023purification} or adversarial rewriting~\cite{gupta2023rewrite}, and detection-based methods such as TextGuard~\cite{pei2024textguard}. Our pattern-informed defense (Section~\ref{sec:defense}) belongs to the input purification paradigm but specifically targets sentence-level complexity and hedging patterns, unlike general-purpose methods designed for token-level perturbations.

\section{Attack Formulation}
\label{sec:formulation}

We formalize the problem of exploiting black-box NLP pipelines through adversarial rewriting under budget-constrained queries. We consider three entities: (1) the \emph{adversary}, an automated agent that generates adversarial rewrites to induce misclassification; (2) the \emph{target pipeline}, a black-box multi-component system that processes natural language inputs through multiple stages to produce a binary decision; and (3) the \emph{end users} who rely on the pipeline's outputs for decision-making.

A claim $x$ is a short text span (typically 1 to 2 sentences) that makes a specific assertion verifiable against evidence. Let $f(x) \in \{0,1\}$ represent the target pipeline that evaluates claims against evidence to output a binary decision, where 1 denotes true claims and 0 denotes false claims. Let $y^*$ be the ground truth label for claim $x$.

The adversary seeks to generate a semantically equivalent rewrite $\tilde{x}$ of the original input $x$ that causes the target pipeline to produce an incorrect output, formally expressed as $f(\tilde{x}) \neq y^*$. The adversary operates under strict black-box conditions, with access only to the pipeline's binary decisions without visibility into model architecture, gradients, training data, confidence scores, or logits. Crucially, multi-stage pipelines may fail at different processing stages (some at evidence retrieval, others at inferential comparison), but the adversary cannot observe which stage fails, making the optimization problem strictly harder than attacking any individual component. Furthermore, the adversary faces a limited query budget due to rate limitations and costs associated with API calls to production systems.

For the generated rewrites to constitute valid adversarial inputs, they must satisfy two constraints. The first is \textit{semantic equivalence}, ensuring the core meaning and intent of the original input remains intact despite the transformation. The second is \textit{textual coherence}, ensuring that the adversarial rewrites maintain grammatical correctness and readability, avoiding artifacts that would distinguish them from legitimate inputs.

Unlike theoretical settings where unlimited queries are possible, we consider realistic production environments where pipelines restrict the number of queries from a single source. This models the practical constraints of attacking deployed systems: commercial APIs enforce query limits, implement rate limiting, and charge per call, making extensive trial-and-error prohibitively expensive.

\paragraph{Adversary Knowledge Model}
The adversary knows the ground truth label $y^*$ for each claim $x$. This assumption is operationally realistic: a disinformation actor submitting a false claim knows it is false and aims for misclassification as true; symmetrically, a propagandist suppressing a true narrative seeks a false classification. No access to internal reasoning or confidence scores is required. We evaluate attacks in both directions (false$\to$true and true$\to$false) across the full dataset.

\section{Methodology}
\label{sec:methodology}

\subsection{Two-Agent Architecture}
\label{sec:two_agent}

Our framework is an agentic LLM system for exploiting black-box NLP pipelines, built on two collaborating agents. An Attacker Agent generates adversarial rewrites and a Prompt Optimization Agent refines the attack strategy based on evaluation feedback.

The Attacker Agent receives the original input $x$ along with system instructions $\wp^{(t)}$ at iteration $t$, and generates an adversarial rewrite:
\begin{equation}
\tilde{x}^{(t)} \sim \mathbb{P}_\mathcal{M}(\cdot|\wp^{(t)} \oplus x)
\end{equation}
where $\oplus$ denotes concatenation and $\mathbb{P}_\mathcal{M}(\cdot|p)$ represents the probability distribution over model $\mathcal{M}$'s outputs conditioned on prompt $p$. The system instructions $\wp^{(t)}$ are high-level directives specifying the agent's role, constraints, and style guidelines for rewriting.

The Prompt Optimization Agent analyzes the outcomes of each attempt and generates refined system instructions for the next iteration:
\begin{equation}
\wp^{(t+1)} \sim \mathbb{P}_\mathcal{M}(\cdot|\wp^{(t)} \oplus x \oplus \tilde{x}^{(t)} \oplus c^{(t)})
\end{equation}
where $c^{(t)}$ denotes the outcomes and scores from the constraint validation module for the rewrite $\tilde{x}^{(t)}$.

\subsection{Rewriting Constraints}
\label{sec:constraints}

Each adversarial rewrite must satisfy two categories of constraints to constitute a valid attack. The first ensures \textit{semantic equivalence}: the rewrite must preserve the original input's meaning as measured by semantic similarity and a binary core assertion check:
\begin{equation}
S(\tilde{x}^{(t)}, x) \geq \tau_S
\end{equation}
\begin{equation}
\text{Sem}(\tilde{x}^{(t)}, x) = 1
\end{equation}
where $S(\cdot,\cdot)$ represents semantic similarity measures with threshold $\tau_S$, and $\text{Sem}(\cdot,\cdot)$ is a binary function confirming the core assertion remains unchanged.

The second ensures \textit{linguistic coherence}:
\begin{equation}
\text{Coh}(\tilde{x}^{(t)}) = 1
\end{equation}
where $\text{Coh}(\cdot)$ evaluates grammatical correctness, readability, and natural language flow, outputting 1 for coherent text and 0 otherwise.

A successful attack is achieved when the adversarial rewrite causes misclassification:
\begin{equation}
f\!\left(\tilde{x}^{(t)}\right) \neq y^*
\end{equation}

All four conditions (3) to (6) must hold simultaneously for a valid adversarial rewrite. Together, they form the \textbf{constraint validation module}.

\subsection{Iterative Optimization}
\label{sec:iterative}

The framework iterates until either all constraints are satisfied or the query budget $T$ is exhausted. We consider two variants of the Prompt Optimization Agent's context.

The \textit{Full-History} variant provides the Prompt Optimization Agent with all previous attempts and their evaluation outcomes:
\begin{equation}
\wp^{(t+1)} \sim \mathbb{P}_\mathcal{M}\!\left(\cdot\,\Big|\,\wp^{(t)} \oplus x \oplus \{\tilde{x}^{(i)}, c^{(i)}\}_{i=1}^{t}\right)
\end{equation}
This enables the agent to identify patterns across attempts and discover target-specific exploitation strategies. This capability becomes increasingly important as pipeline robustness increases and single-attempt evasion attacks become less likely to succeed, though it comes at the cost of longer prompts.

The \textit{Previous-Only} variant provides only the most recent attempt:
\begin{equation}
\wp^{(t+1)} \sim \mathbb{P}_\mathcal{M}(\cdot|\wp^{(t)} \oplus x \oplus \tilde{x}^{(t)} \oplus c^{(t)})
\end{equation}
This reduces context length and enables faster iteration.

To encourage exploration diversity, the Attacker Agent uses a temperature schedule: 1.0 initially, increasing by $+0.1$ per iteration after iteration 5, capped at 1.5. This schedule was calibrated on a held-out set of 50 claims; alternative schedules (linear increase from 1.0, fixed 1.2) yielded comparable but slightly lower success, suggesting robustness to moderate variation. The Prompt Optimization Agent maintains constant temperature 1.0 throughout.

\paragraph{Two-Phase Search Intuition}
The two-agent loop implicitly implements exploration-exploitation. Early iterations generate diverse rewrites that broadly probe the decision boundary. A rewrite that satisfies semantic constraints but fails to flip the classification serves as a \emph{warm-start point}, indicating that a nearby region of the perturbation space may contain valid adversarial examples. The Prompt Optimization Agent then shifts from exploration to exploitation: analyzing which constraints were satisfied and which failed across the history, it concentrates subsequent perturbation on the dimensions most likely to induce boundary crossing. The temperature schedule reinforces this transition: conservative sampling ($\tau{=}1.0$) supports early exploration, while increased temperature ($\tau {\to} 1.5$) provides diversity within the narrowed search region.

\begin{figure*}[!t]
\centering
\includegraphics[width=0.95\textwidth]{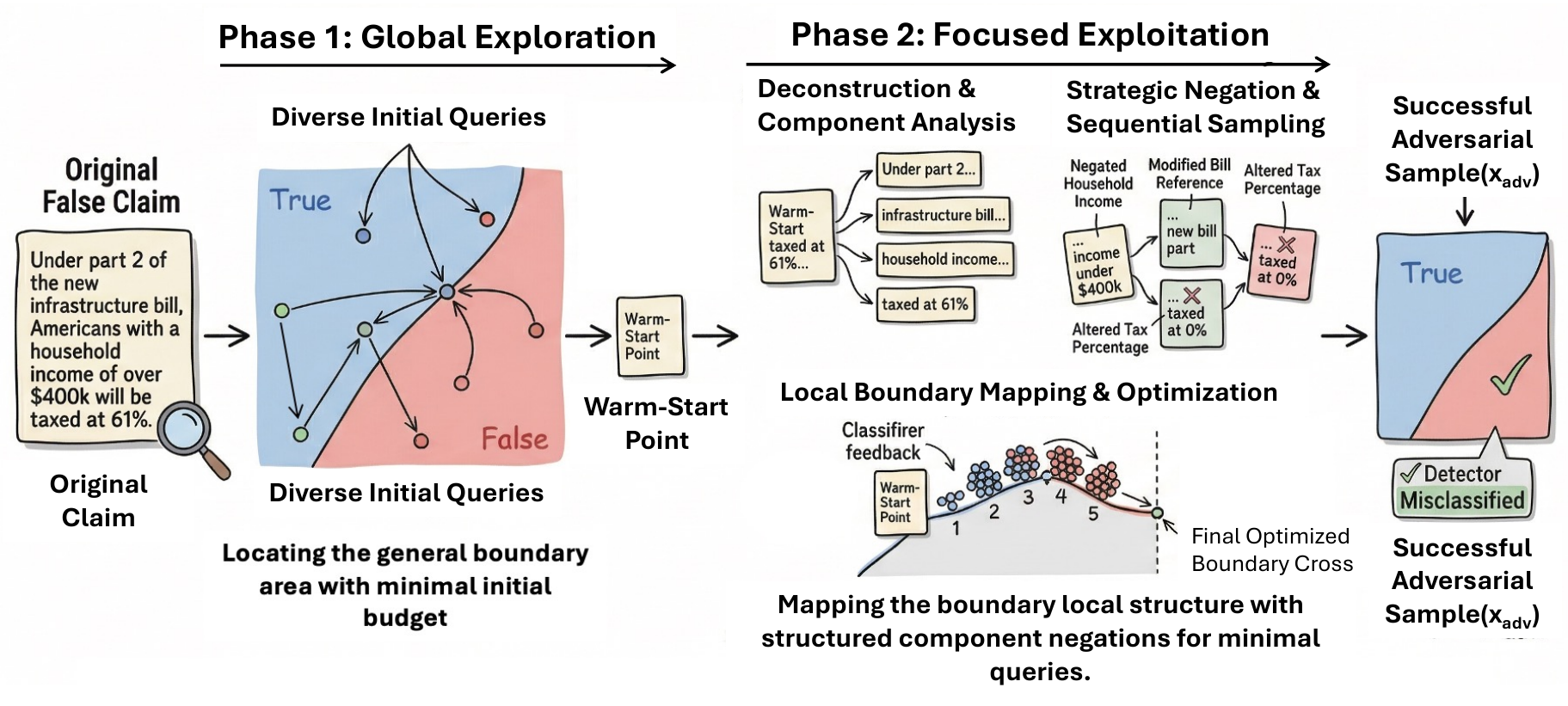}
\caption{Detailed system architecture of the agentic adversarial rewriting framework. The diagram specifies each component of the constraint validation module (MPNet similarity, BERTScore, GPT-4o semantic equivalence check, and GPT-4o-mini coherence check), the feedback loop between the target detector and the Prompt Optimization Agent, and the decision logic governing iteration termination. The right panel traces a complete attack trajectory through three iterations, illustrating how constraint and detector feedback jointly guide the Prompt Optimization Agent toward a successful adversarial rewrite.}
\label{fig:methodology}
\end{figure*}

The complete procedure is detailed in Algorithm~\ref{alg:attack} and the full system architecture is presented in Fig.~\ref{fig:methodology}.

\begin{algorithm}[t]
\caption{Agentic Adversarial Rewriting Attack}
\label{alg:attack}
\begin{algorithmic}[1]
    \Require Original claim $x$, black-box detector $f$, ground truth $y^*$, max iterations $T$
    \Ensure Adversarial rewriting $\tilde{x}$ that preserves semantic meaning and coherence
    \State $t \leftarrow 0$
    \State $\text{instructions} \leftarrow \text{InitialSystemInstructions}()$
    \State $\text{history} \leftarrow \emptyset$
    \While{$t < T$}
        \State $t \leftarrow t + 1$
        \State $\tilde{x}^{(t)} \leftarrow \text{AttackerAgent}(x, \text{instructions})$
        \State $\text{semantic\_sim} \leftarrow S(\tilde{x}^{(t)}, x) \geq \tau_S$
        \State $\text{sem\_check} \leftarrow \text{Sem}(\tilde{x}^{(t)}, x) = 1$
        \State $\text{coherence} \leftarrow \text{Coh}(\tilde{x}^{(t)}) = 1$
        \State $\text{attack\_success} \leftarrow f(\tilde{x}^{(t)}) \neq y^*$
        \State $\text{all\_satisfied} \leftarrow \text{semantic\_sim} \wedge \text{sem\_check} \wedge \text{coherence} \wedge \text{attack\_success}$
        \If{all\_satisfied}
            \Return $\tilde{x}^{(t)}$
        \EndIf
        \State $\text{history} \leftarrow \text{history} \cup \{(\tilde{x}^{(t)}, \text{checks})\}$
        \If{UseFullHistory}
            \State $\text{instructions} \leftarrow \text{PromptOptAgent}(x, \text{history})$
        \Else
            \State $\text{instructions} \leftarrow \text{PromptOptAgent}(x, \{(\tilde{x}^{(t)}, \text{checks})\})$
        \EndIf
    \EndWhile
    \State \Return \texttt{NULL}
\end{algorithmic}
\end{algorithm}

\subsection{Semantic Equivalence and Linguistic Coherence Measurements}
\label{sec:measurements}

The constraint validation module enforces four simultaneous checks on every adversarial rewrite.

\paragraph{MPNet Embedding Similarity}
We compute cosine similarity using the \texttt{sentence-transformers/all-mpnet-base-v2} model~\cite{reimers2019sentence} via the HuggingFace SentenceTransformers library~\cite{huggingface2023sentence}:
\begin{equation}
S_{\text{MPNet}}(\tilde{x}^{(t)}, x) \geq \tau_{\text{MPNet}}
\end{equation}
with threshold $\tau_{\text{MPNet}} = 0.61$.

\paragraph{BERTScore Similarity}
We use BERTScore with the \texttt{microsoft/deberta-xlarge-mnli} model~\cite{he2021deberta,huggingface2021deberta}, computing the F1 score with baseline rescaling:
\begin{equation}
S_{\text{BERT}}(\tilde{x}^{(t)}, x) \geq \tau_{\text{BERT}}
\end{equation}
with threshold $\tau_{\text{BERT}} = 0.77$.

\paragraph{LLM-Based Semantic Check}
A GPT-4o-based binary verification confirms that the core assertion remains unchanged, guarding against subtle additions or omissions of essential facts:
\begin{equation}
\text{GPT4-Eq}(\tilde{x}^{(t)}, x) = 1
\end{equation}
This check uses few-shot examples to ensure consistent evaluation. We acknowledge that using an LLM to validate LLM-generated rewrites introduces a potential same-family bias; we mitigate this through two independent embedding-based checks (MPNet and BERTScore) that provide cross-architecture validation.

\paragraph{Linguistic Coherence}
A GPT-4o-mini-based evaluator~\cite{openai2024gptmini} assesses grammatical correctness, readability, and natural language flow:
\begin{equation}
\text{GPT4-Coh}(\tilde{x}^{(t)}) = 1
\end{equation}

\paragraph{Threshold Derivation}
The thresholds $\tau_{\text{MPNet}}$ and $\tau_{\text{BERT}}$ were derived through a human evaluation study with three annotators on 200 original/rewritten claim pairs from a validation subset of LIAR-New. We employed bin-based stratified sampling across MPNet and BERTScore quartiles and selected thresholds to minimize false positives. Inter-annotator agreement measured by Fleiss' Kappa was 1.0 for coherence and 0.610 for semantic equivalence~\cite{landis1977measurement}. We report the scope and limitations of this calibration study in Section~\ref{sec:limitations}.

\section{Experiments}
\label{sec:experiments}

We evaluate our framework on evidence-based misinformation detection, a domain that exhibits the canonical multi-stage pipeline structure: natural language input, natural language evidence retrieval, and natural language inferential comparison.

\subsection{Data}

We use the LIAR-New dataset~\cite{pelrine2023towards}, containing 1,957 real-world political statements from PolitiFact categorized into three verifiability levels: \textit{Possible} (927), \textit{Hard} (581), and \textit{Impossible} (449). We retain only Possible and Hard categories (objectively determinable veracity). The six-class labels from the original LIAR dataset~\cite{wang2017liar} are mapped to binary: \textbf{False} (Pants-fire + False + Mostly-false = 1,663) and \textbf{True} (Half-true + Mostly-true + True = 294), following the convention where ``Mostly-false''/``Half-true'' represents the natural decision threshold. We randomly selected a stratified subset of 500 samples, preserving class distribution. Although LIAR-New consists of political statements, the claims span diverse policy domains (healthcare, economy, immigration, foreign policy, criminal justice).

\subsection{Implementation}

\subsubsection{Target Pipelines}

We evaluated against four evidence-based misinformation detection pipelines, each exhibiting the canonical three-stage structure (input intake, evidence retrieval, inferential comparison).

\textbf{Verifact}~\cite{tian2024web} employs a two-agent architecture with Google Search API retrieval and GPT-4o-mini~\cite{openai2024gptmini} (substituting GPT-4-0125 for cost). \textbf{ICL}~\cite{singhal2024evidence} uses in-context learning with Google Search API and Qwen2-VL-72B-Instruct~\cite{yang2024qwen2,qwen2024qwen2vl}, outputting binary decisions. \textbf{ClaimBuster}~\cite{hassan2017claimbuster} is a legacy system whose \texttt{fact\_matcher} API~\cite{idirlab2017claimbuster} relies on lexical retrieval; \texttt{not\_enough\_info} responses were treated as incorrect. We include it to anchor the vulnerability spectrum. \textbf{Perplexity} uses the Sonar model~\cite{perplexity2025models} with real-time web search, achieving performance comparable to GPT-4o-mini~\cite{perplexityteam2024}.

\subsubsection{Attack Methods}

\paragraph{Our Framework}
Both the Attacker Agent and Prompt Optimization Agent utilized GPT-4o. The temperature schedule is described in Section~\ref{sec:iterative}. We evaluated three variants: \textit{Full History}, where the Prompt Optimization Agent receives all previous attempts; \textit{Previous Step}, where it receives only the most recent attempt; and \textit{Attacker Only}, where no prompt optimization occurs and the Attacker Agent's prompt is never updated. All variants operated under a maximum budget of 10 queries per input.

\paragraph{Baseline Methods}
We implemented four token-level baselines via TextAttack~\cite{morris2020textattack}: \textbf{CLARE}~\cite{li2021contextualized} (mask-then-infill), \textbf{DeepWordBug}~\cite{gao2018black} (character-level transformations), \textbf{TextBugger}~\cite{li2019textbugger} (hybrid character/word perturbations), and \textbf{TextFooler}~\cite{jin2020bert} (synonym substitution guided by prediction probabilities). All used a surrogate model: RoBERTa-base~\cite{liu2019roberta,huggingface2024roberta} fine-tuned on LIAR~\cite{wang2017liar} with class-weighted loss and maximum sequence length of 128. All methods were allowed 10 queries and evaluated under the same semantic equivalence and coherence constraints (Sections~\ref{sec:constraints} and~\ref{sec:measurements}).

\paragraph{Notes on Baseline Comparison}
We exclude recent LLM-based adversarial methods (GCG~\cite{zou2023universal}, PromptAttack~\cite{xu2024promptattack}, BEAST~\cite{sadasivan2024beast}) because they target jailbreaking of individual LLMs, not multi-stage pipeline evasion, and require thousands of optimization steps without semantic equivalence constraints. The comparison with token-level baselines is structurally asymmetric: our framework queries the target pipeline directly, while baselines must query a surrogate model because they require prediction probabilities unavailable from the target. The Attacker Only variant isolates the contribution of prompt optimization from the advantage of direct target access.

\subsection{Target Pipeline Baseline Performance}

\begin{table}[t]
\centering
\caption{Baseline performance of four NLP pipelines for misinformation detection on binary classification (True/False) using the LIAR-New dataset. Perplexity achieves the highest performance across all metrics.}
\label{tab:baseline}
\resizebox{\columnwidth}{!}{%
\begin{tabular}{lcccc}
\toprule
\textbf{Metrics} & \textbf{ICL} & \textbf{Verifact} & \textbf{ClaimBuster} & \textbf{Perplexity} \\
\midrule
Accuracy       & 71.20 & 82.40 & 61.00 & \textbf{86.20} \\
Macro F1       & 55.00 & 65.11 & 49.89 & \textbf{68.88} \\
Macro Recall   & 56.81 & 64.49 & 40.12 & \textbf{66.22} \\
Macro Precision& 54.86 & 65.85 & 65.96 & \textbf{74.16} \\
\bottomrule
\end{tabular}}
\end{table}

Table~\ref{tab:baseline} presents baseline performance. Perplexity achieves the highest accuracy (86.20\%) and Macro F1 (68.88\%), followed by Verifact (82.40\%), ICL (71.20\%), and ClaimBuster (61.00\%). The accuracy--Macro F1 gap across all systems reflects class imbalance (1,663 false vs.\ 294 true claims). ClaimBuster's high precision (65.96\%) but low recall (40.12\%) indicates it correctly identifies false claims but misses many true claims.

\subsection{Attack Success Results}

\begin{table}[t]
\centering
\caption{Attack success rates (\%) against four NLP pipelines. 95\% Wilson confidence intervals (CI) are reported for our Full History variant; all CIs are $\pm$1.5 to 4.3\,pp (see text). Baseline methods use a RoBERTa surrogate model for prediction probabilities.$^\dagger$}
\label{tab:attack_success}
\resizebox{\columnwidth}{!}{%
\begin{tabular}{lcccc}
\toprule
\textbf{Method} & \textbf{ICL} & \textbf{Verifact} & \textbf{ClaimBuster} & \textbf{Perplexity} \\
\midrule
CLARE$^\dagger$             & 3.90  & 1.46  & 26.56 & 0.46 \\
DeepWordBug$^\dagger$       & 1.97  & 1.46  & 10.82 & 0.23 \\
TextBugger$^\dagger$        & 2.80  & 1.94  & 22.95 & 0.00 \\
TextFooler$^\dagger$        & 3.09  & 2.18  & 23.93 & 1.16 \\
\midrule
Ours (Attacker Only)    & 20.11 & 35.92 & 95.02 & 14.55 \\
Ours (Previous Step)    & 25.00 & 39.56 & 93.05 & 18.14 \\
Ours (Full History)     & \textbf{30.35} & \textbf{40.34} & \textbf{97.02} & \textbf{19.95} \\
\bottomrule
\end{tabular}}
\end{table}

Table~\ref{tab:attack_success} presents attack success rates across all four target pipelines. We focus the analysis on the three modern systems (ICL, Verifact, Perplexity), as these represent the pipeline architectures currently deployed in practice. ClaimBuster results are reported for completeness as a legacy anchor point.

\paragraph{Modern Pipeline Results} Our Full History variant achieved 40.34\% [$\pm$4.3\,pp] against Verifact, 30.35\% [$\pm$4.0\,pp] against ICL, and 19.95\% [$\pm$3.5\,pp] against Perplexity (95\% Wilson CIs, $n{=}500$). The strongest token-level baseline (CLARE) achieved at most 3.90\% [$\pm$1.7\,pp] on ICL and below 2.18\% on other modern targets. However, this comparison is structurally asymmetric: our framework queries the target pipeline directly, while baselines query a surrogate model because they require prediction probabilities unavailable from the target. To disentangle the effect of direct target access from that of semantic-level rewriting, we compare against our Attacker Only variant, which shares the same direct target access but performs no iterative optimization. Attacker Only achieves 14.55 to 35.92\% on modern targets, confirming that the majority of the gap over token-level baselines stems from semantic-level rewriting rather than from the optimization loop or the access advantage alone.

\paragraph{Legacy Baseline} Against ClaimBuster, our framework achieved 97.02\% [$\pm$1.5\,pp], compared to CLARE's 26.56\% [$\pm$3.9\,pp]. ClaimBuster's near-complete exploitability stems from its legacy architecture: its lexical retrieval stage frequently returns \texttt{not\_enough\_info} for reformulated claims, reflecting brittle keyword matching rather than a sophisticated adversarial challenge. We treat this result as an architectural lower bound rather than a primary finding.

\begin{figure*}[!t]
\centering
\includegraphics[width=0.8\textwidth]{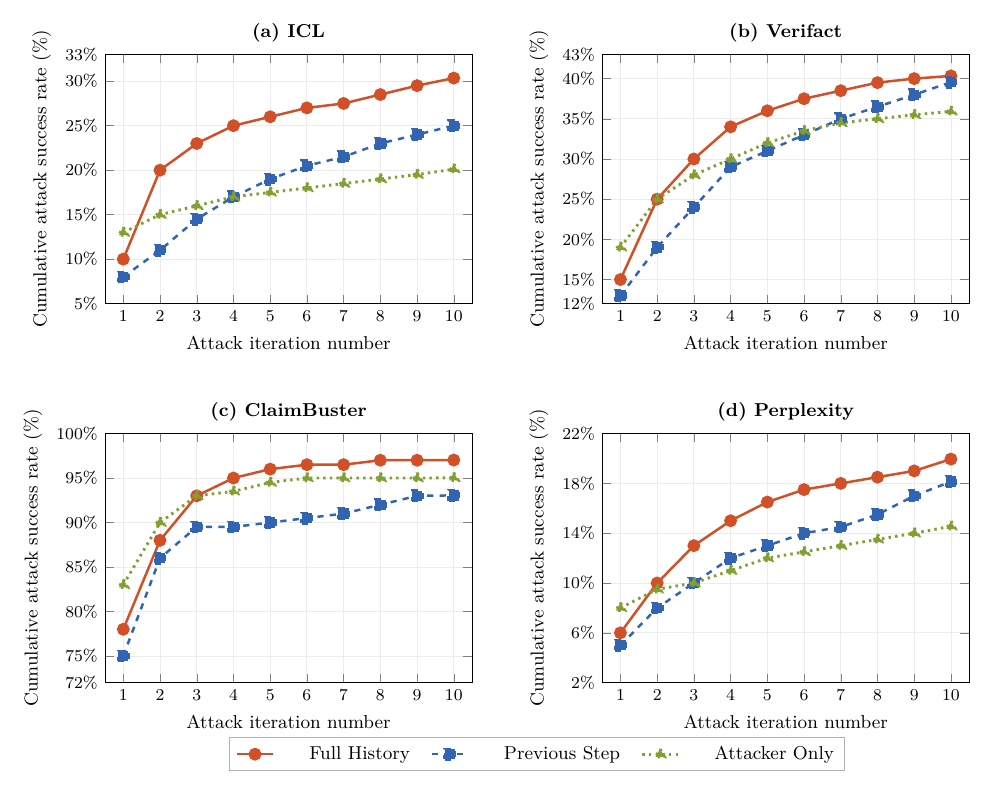}
\caption{Cumulative attack success rates (\%) for three framework variants over 10 attack iterations on four target pipelines: (a) ICL, (b) Verifact, (c) ClaimBuster, and (d) Perplexity. Full History consistently achieves the highest success rate, with the largest marginal gains on the most robust targets.}
\label{fig:iterations}
\end{figure*}

\paragraph{Prompt Optimization Value} Among our variants, Full History consistently outperformed Previous Step, which in turn outperformed Attacker Only (Fig.~\ref{fig:iterations}). The marginal value of prompt optimization scales inversely with target vulnerability: the Prompt Optimization Agent adds only 2.00\,pp against ClaimBuster (95.02\% $\to$ 97.02\%), where any valid rewrite suffices, but contributes 10.24\,pp against ICL (20.11\% $\to$ 30.35\%), where the search problem is harder and history-conditioned strategy refinement is most valuable. This pattern is consistent with the theoretical expectation that optimization contributes most when the feasible region of the constrained problem is narrow.

\paragraph{Statistical Notes} All confidence intervals are 95\% Wilson binomial intervals ($n{=}500$). No baseline CI overlaps with our Full History CI on any modern target, confirming statistical significance. These intervals capture sampling variability but not LLM generation stochasticity (Section~\ref{sec:limitations}).

\paragraph{Vulnerability Spectrum} The results reveal a vulnerability spectrum associated with three architectural properties: (1)~\emph{evidence retrieval mechanism} (lexical matching is catastrophically vulnerable; real-time search provides resilience), (2)~\emph{retrieval-inference coupling} (tight integration resists inter-stage mismatch), and (3)~\emph{baseline accuracy} (higher accuracy widens the decision margin). Perplexity (19.95\%) proves most robust, ICL (30.35\%) and Verifact (40.34\%) occupy the middle range, and ClaimBuster (97.02\%) anchors the vulnerable end.

\subsection{Exploitation Pattern Analysis}

\begin{table*}[t]
\centering
\caption{Text analysis of successful adversarial rewrites (Ours, Full History) compared to original inputs. Metrics span semantic similarity, lexical changes, syntactic/structural modifications, and stylistic differences.}
\label{tab:text_analysis}
\begin{tabular}{lcccccc}
\toprule
\textbf{Text} & \textbf{BERTScore} & \textbf{Levenshtein} & \textbf{Parse Tree Dist.} & \textbf{Text Length} & \textbf{Perplexity} & \textbf{Flesch RE} \\
\midrule
Original    & --     & --      & --     & 103.29 & 207.17 & 64.40 \\
ICL         & 0.8394 & 85.7204 & 0.3287 & 159.46 & 73.06  & 39.91 \\
Verifact    & 0.8382 & 85.4274 & 0.3212 & 154.68 & 101.34 & 39.58 \\
ClaimBuster & 0.8550 & 69.5565 & 0.3050 & 132.34 & 110.56 & 43.48 \\
Perplexity  & 0.8376 & 82.6296 & 0.3265 & 148.34 & 88.84  & 43.28 \\
\bottomrule
\end{tabular}
\end{table*}

We analyzed the characteristics of successful Full History adversarial rewrites across all four targets (Table~\ref{tab:text_analysis}). The analysis reveals four recurring patterns in successful adversarial rewrites. We describe each pattern below and note the pipeline stages where they are most likely to have an effect, though we acknowledge that these stage attributions are correlational (based on the nature of the transformation and which pipelines are most affected) rather than causally verified through controlled ablation.

\paragraph{Pattern 1: Hedging and Ambiguity Injection} Term frequency--inverse document frequency (TF-IDF) analysis reveals a systematic vocabulary shift: original texts frequently use direct terms such as ``says,'' ``and,'' ``are,'' ``people,'' and ``because,'' while successful adversarial rewrites employ hedging terms including ``that,'' ``as,'' ``might,'' ``some,'' ``reportedly,'' and ``potentially.'' This hedging introduces ambiguity that is consistent with degradation of the evidence retrieval stage, making it harder for the pipeline to match claims against definitive evidence.

\paragraph{Pattern 2: Structural Elaboration} Our framework consistently generated longer texts, with character counts increasing from 103.29 (original) to 132.34 to 159.46 (adversarial), representing a 28 to 54\% increase. This elaboration dilutes the core claim signal across a longer text span, which is consistent with challenging the retrieval mechanism's ability to extract the key verifiable assertion.

\paragraph{Pattern 3: Complexity Escalation} The Flesch Reading Ease score dropped from 64.40 (approximately 8th-grade level) to 39.58 to 43.48 (college level), a 32 to 39\% decrease. Adversarial rewrites also exhibited lower perplexity (73.06 to 110.56) compared to originals (207.17), consistent with the observation that LLM-generated text is more predictable to language models~\cite{holtzman2020curious}. This complexity escalation is consistent with disrupting the inferential comparison stage, where the mismatch between complex adversarial inputs and simpler retrieved evidence may degrade entailment judgments.

\paragraph{Pattern 4: Syntactic Restructuring} Parse tree distances of 0.3050 to 0.3287, combined with high Levenshtein distances (69.56 to 85.72), confirm substantial structural and lexical modification despite meaning preservation. BERTScore values (0.8376 to 0.8550) remained well above the threshold of 0.77, confirming semantic fidelity. These modifications are consistent with disrupting pattern-matching heuristics at the retrieval stage while preserving the underlying assertion.

These stage-level associations are consistent with the vulnerability spectrum: ClaimBuster is maximally vulnerable to retrieval-associated patterns (1, 2, 4), while Perplexity's real-time search neutralizes retrieval-stage disruption. Controlled ablation (e.g., constraining the rewriter to a single pattern) is needed to establish causal attribution.

\subsection{Pattern-Informed Defense}
\label{sec:defense}

\begin{table}[t]
\centering
\caption{Effectiveness of text simplification defense against our Full History variant across four NLP pipelines for misinformation detection.}
\label{tab:defense}
\resizebox{\columnwidth}{!}{%
\begin{tabular}{lccccc}
\toprule
\textbf{Pipeline} & \multicolumn{2}{c}{\textbf{Flesch RE}} & \multicolumn{3}{c}{\textbf{Attack Success Rate (\%)}} \\
\cmidrule(lr){2-3}\cmidrule(lr){4-6}
 & Attack & Defense & Attack & Defense & Reduction \\
\midrule
ICL          & 39.91 & 52.91 & 30.35 & 10.57 & \textbf{65.18} \\
Verifact     & 39.58 & 53.31 & 40.34 & 20.77 & 48.50 \\
ClaimBuster  & 43.48 & 55.67 & 97.02 & 58.28 & 39.93 \\
Perplexity   & 43.28 & 54.96 & 19.95 & 10.43 & 47.73 \\
\bottomrule
\end{tabular}}
\end{table}

Based on the exploitation patterns identified above, particularly complexity escalation (Pattern~3), we developed a defense through input transformation via \textit{text simplification} as a preprocessing step. We employed a GPT-4o-based text simplifier instructed to: (1) preserve exact meaning and factual content, (2) eliminate unnecessary complexity and ambiguous qualifiers, (3) enhance clarity through simple, direct language, and (4) maintain natural language flow. This defense targets the linguistic properties that our framework exploits by normalizing the complexity of inputs before they enter the pipeline.

Table~\ref{tab:defense} presents the defense results. The text simplification defense increased the Flesch Reading Ease score by 10 to 14 points across all adversarial rewrites, bringing them closer to the readability level of original claims. On modern pipelines, ICL benefited most from the defense, with a 65.18\% reduction in attack success rate (30.35\% to 10.57\%). Verifact saw a 48.50\% reduction (40.34\% to 20.77\%), and Perplexity achieved a 47.73\% reduction (19.95\% to 10.43\%). ClaimBuster, while experiencing a 39.93\% reduction, retained the highest post-defense attack success rate at 58.28\%.

The defense effectiveness varies with vulnerability type: ICL benefits most (65.18\% reduction) because its exploitation is primarily associated with complexity escalation at the inference stage, while ClaimBuster retains 58.28\% post-defense attack success because its vulnerability is architectural (brittle lexical retrieval) rather than pattern-driven. This suggests that text simplification addresses inference-stage complexity mismatch but cannot compensate for fundamental retrieval-stage weaknesses. A comprehensive defense evaluation incorporating ensemble defenses, perturbation detection, and consistency checking is important future work.

\subsection{Cost Analysis}
\label{sec:cost}

Our framework costs approximately \$0.30 to \$0.50 per claim (GPT-4o at \$5/\$15 per 1M input/output tokens), yielding \$150 to \$250 for the 500-claim experiment per target pipeline. Average wall-clock time is 45 to 90 seconds per claim for Full History, reduced to approximately 30 seconds for highly vulnerable targets via early termination. The convergence curves in Fig.~\ref{fig:iterations} show that Full History achieves 60 to 75\% of its final attack success within 4 iterations, indicating that a budget of $T{=}5$ would capture the majority of exploitable claims at half the cost, while extending beyond $T{=}10$ would yield minimal additional gains ($<$2\,pp).

\section{Discussion}

\paragraph{Why Agentic Optimization, Not Reinforcement Learning}
A natural question is why we use an agentic two-agent approach rather than reinforcement learning methods such as PPO~\cite{schulman2017proximal}. Two factors motivated this choice. First, RL requires dense reward signals, but our setting provides only binary feedback with no directional information. Sparse binary rewards yield extremely high variance in policy gradient estimates~\cite{schulman2017proximal}, and exploration under such sparsity may require orders of magnitude more iterations than our 10-query budget permits. Second, RL requires exemplar trajectories to bootstrap learning; in our setting, such exemplars would be the successful rewrites we aim to generate, creating a circular dependency.

Our framework sidesteps both limitations by leveraging the pretrained linguistic knowledge of LLMs to generate promising adversarial candidates from the outset. The Prompt Optimization Agent then refines strategy using binary feedback combined with iteration history, achieving an efficiency that learning-based algorithms cannot match within a 10-query budget.

\paragraph{Why Two Agents, Not One}
A simpler alternative would be a single LLM with self-reflection. The Attacker Only variant (Table~\ref{tab:attack_success}) tests exactly this: a single agent generating rewrites without iterative optimization, achieving 14.55 to 35.92\% on modern targets. The full framework adds 5.35 to 10.24\,pp over this baseline, with the largest gains on the most robust targets. This confirms that semantic-level rewriting is the primary contributor, while the Prompt Optimization Agent provides essential marginal gains against robust pipelines where single-attempt attacks fail. The two-agent decomposition is therefore the minimum complexity required for effective boundary search across the full vulnerability spectrum.

\paragraph{Architectural Determinants of Vulnerability}
Pipeline vulnerability forms a spectrum determined by three properties: evidence retrieval mechanism, retrieval-inference coupling, and baseline accuracy. Legacy systems with lexical retrieval are near-completely exploitable (97.02\%) because keyword matching is trivially defeated. Modern LLM-based pipelines resist retrieval-stage attacks but remain vulnerable at inference, where complexity escalation creates stylistic mismatch with retrieved evidence. Pipelines with real-time evidence grounding resist both stages most effectively (19.95\%). This analysis generalizes: any multi-stage system relying on surface-level lexical matching can be expected to exhibit similar vulnerability. The finding that modern pipelines are not immune (30 to 40\% attack success without gradient access) underscores that semantic retrieval does not fully protect the inference stage.

\paragraph{Query Efficiency Through Implicit Boundary Search}
Each adversarial rewrite probes the target's decision space within a feasible region $\mathcal{B}(x, \boldsymbol{\tau})$ defined by the intersection of similarity constraints (MPNet, BERTScore, LLM-based check). The two-agent framework implicitly decomposes search into exploration (diverse rewrites locating the boundary) followed by exploitation (history-informed refinement). This explains why the Prompt Optimization Agent's marginal contribution scales inversely with vulnerability: against ClaimBuster the boundary intersects a large portion of $\mathcal{B}(x, \boldsymbol{\tau})$ (+2.00\,pp), while against ICL the feasible region is narrow and exploitation is essential (+10.24\,pp). Full History outperforms Previous Step because accumulated observations enable finer boundary localization, and achieves steep early gains followed by diminishing returns (Fig.~\ref{fig:iterations}), consistent with exploration-exploitation transition.

\paragraph{Validity of Semantic Equivalence Evaluation}
Our constraint validation module uses three independent methods: two embedding-based similarity measures (MPNet and BERTScore) and one LLM-based binary verification (GPT-4o). Thresholds were calibrated against human judgments on 200 pairs with Fleiss' Kappa of 1.0 (coherence) and 0.610 (semantic equivalence). We acknowledge three limitations: (1)~the LLM-based check uses a model from the same family as the generator, introducing potential same-family bias, mitigated by the cross-architecture embedding checks; (2)~the calibration used validation pairs, not actual attack outputs; and (3)~the Kappa of 0.610, while substantial, reflects inherent ambiguity in semantic equivalence judgments for adversarial rewrites. Successful rewrites maintain BERTScore values of 0.8376 to 0.8550, well above the threshold, and the four identified patterns are stylistic rather than semantic transformations. Human evaluation of actual attack outputs remains important future work.

\paragraph{Generalizability and Model Selection}
While we instantiate our framework on misinformation detection, the approach is domain-agnostic. Any system that (a)~accepts natural language input, (b)~processes it through multiple NL stages, and (c)~returns a discrete decision is a candidate for adversarial exploitation. The two-agent architecture and prompt optimization mechanism require no domain-specific adaptation; only the rewriting constraints may need recalibration. We selected GPT-4o as the attack model for its strong instruction-following and generation quality; the framework architecture is model-agnostic and could use any sufficiently capable LLM. GPT-4o-mini serves as the coherence evaluator because coherence checking is a simpler task that does not require the full model's capabilities, reducing cost without sacrificing evaluation quality.

\section{Limitations}
\label{sec:limitations}

While our framework demonstrates effective adversarial exploitation of black-box NLP pipelines, several limitations warrant acknowledgment.

First, the semantic equivalence validation relies in part on a GPT-4o-based binary check from the same model family as the Attacker Agent. Although two independent embedding-based measures (MPNet and BERTScore) provide cross-architecture validation, and thresholds were calibrated against human judgments on 200 pairs with Fleiss' Kappa of 1.0 (coherence) and 0.610 (semantic equivalence), the human evaluation did not cover actual successful adversarial rewrites. A comprehensive human evaluation of attack outputs would quantify the false-positive rate of the constraint validation module.

Second, our evasion rates are point estimates from a single experimental run. The 95\% Wilson confidence intervals capture sampling variability but do not account for the additional stochasticity introduced by LLM generation with temperature-based sampling. Multi-seed evaluation would provide a more complete picture of result stability.

Third, our evaluation uses only the LIAR-New dataset~\cite{pelrine2023towards} of political statements from PolitiFact. Although the claims span diverse policy domains and the attack framework is domain-agnostic, generalization to structurally different datasets such as FEVER~\cite{thorne2018fever} and AVERITEC~\cite{schlichtkrull2023averitec} should be validated empirically. Similarly, evaluation is limited to evidence-based misinformation detection; validation across other multi-stage domains (RAG-based question answering, tool-augmented LLM applications) remains important future work. The vulnerability spectrum we observe appears driven by pipeline architecture rather than claim content, but this hypothesis requires direct experimental confirmation.

Fourth, we do not evaluate cross-pipeline transferability of adversarial rewrites. Determining whether a rewrite crafted to evade one pipeline also evades others would reveal shared versus pipeline-specific vulnerabilities. Additionally, the framework currently targets pipelines with binary outputs; extending to multi-class or free-text output spaces requires adapting the feedback mechanism.

Fifth, the four exploitation patterns are described post-hoc through aggregate statistics and their stage attributions are correlational. Controlled ablation experiments (e.g., constraining the rewriter to produce only hedging-based or only complexity-based rewrites) would establish causal attribution and quantify per-pattern contributions.

Sixth, the framework depends on commercial LLM APIs (GPT-4o) that may change over time. Model updates, safety filter modifications, or API deprecation could affect reproducibility. While the two-agent architecture and optimization strategy are model-agnostic, the specific evasion rates are tied to the model version used at evaluation time.

\section{Ethical Considerations}
\label{sec:ethics}

This work presents a framework for adversarial exploitation of NLP pipelines deployed in misinformation detection. We recognize the dual-use concern and address it through deliberate design and disclosure choices.

\paragraph{Responsible Disclosure}
All target systems are publicly available research prototypes or commercial APIs. Our evaluation used a public benchmark dataset (LIAR-New) of previously published political statements. The adversarial rewrites constitute novel adversarial content (stylistically transformed versions of existing claims) that could, in principle, evade deployed systems. However, they preserve the factual content of the original claims and are therefore no more or less harmful than the original misinformation from which they are derived. No novel misinformation was generated or disseminated.

\paragraph{Defensive Intent}
The primary motivation is to expose vulnerabilities \emph{before} malicious actors discover them independently. The vulnerability spectrum and exploitation pattern analysis provide system designers with actionable intelligence for hardening their pipelines. Our pattern-informed defense demonstrates that adversarial evaluation insights translate directly into protective countermeasures (up to 65.18\% evasion reduction). This research is analogous to penetration testing in network security.

\paragraph{Scope Limitations}
The framework is deliberately constrained: the 10-query budget bounds adversarial cost, strict semantic equivalence constraints prevent transformation of claims into genuinely different (more harmful) content, and we do not release the specific system prompts or provide a turnkey attack tool. The contribution is at the level of the two-phase architecture and boundary-search formulation.

\paragraph{Broader Impact}
As multi-stage NLP systems are deployed in high-stakes domains (medical triage, financial compliance, content moderation), understanding their adversarial robustness becomes a matter of public interest. We advocate for adversarial red-teaming as a standard evaluation component for all multi-stage NLP systems deployed in safety-critical contexts.

\section{Conclusion}

We presented a two-agent framework for exploiting black-box NLP pipelines through adversarial rewriting under budget-constrained queries. Operating with only binary feedback and a 10-query budget, the framework achieves 19.95 to 40.34\% evasion on modern LLM-based misinformation detection pipelines ($\pm$4.3\,pp, 95\% CI), compared to at most 3.90\% for token-level baselines that rely on surrogate models because they cannot operate under our threat model. Even our non-iterative Attacker Only variant, which shares the same direct target access, achieves 14.55 to 35.92\%, confirming that semantic-level rewriting provides a fundamental advantage independent of the optimization strategy.

A legacy system with static lexical retrieval proves near-completely exploitable (97.02\%), anchoring a vulnerability spectrum that is associated with three architectural properties: evidence retrieval mechanism, retrieval-inference coupling, and baseline accuracy. Analysis of successful rewrites reveals four exploitation patterns targeting distinct pipeline stages, and a pattern-informed text simplification defense reduces evasion by up to 65.18\%.

As multi-stage NLP systems are increasingly deployed in high-stakes domains, including fact-checking, medical triage, and financial compliance, principled adversarial evaluation becomes essential. The vulnerability spectrum we identify provides system designers with actionable intelligence for architectural hardening. Future work should validate the framework on additional NLP pipeline domains (RAG-based question answering, tool-augmented LLM applications), conduct controlled ablation experiments to establish causal attribution of individual exploitation patterns, and develop multi-technique defenses combining text simplification with input perturbation detection and consistency checking.

\section*{Supplementary Material}
Supplementary material accompanies this paper and includes: (A)~representative adversarial rewrite examples that illustrate the four exploitation patterns, (B)~human annotation guidelines used in the threshold calibration study, and (C)~a complete walk-through of the iterative prompt optimization process across six iterations.


\bibliographystyle{IEEEtran}
\bibliography{bibliography}


\end{document}